\def\paperID{6797}
\def\confName{CVPR}
\def\confYear{2024}

\def\paperTitle{Extreme Point Supervised Instance Segmentation}

\def\authorBlock{
    Hyeonjun Lee$^{1,3}$ \qquad
    Sehyun Hwang$^{2}$\qquad
    Suha Kwak$^{2,3}$ \\
    $^1$Lunit Inc. \qquad \ \ $^2$Dept. of CSE, POSTECH \qquad \ \   $^3$Graduate School of AI, POSTECH \\
    {
    \tt\small hyeonjun1882@lunit.io,
    \tt\small \{sehyun03, suha.kwak\}@postech.ac.kr}
}

\newif\ifreview 
\newif\ifarxiv 
\newif\ifcamera \newcommand{\cameraready}{\cameratrue}
\newif\ifrebuttal 

\cameraready
\pdfoutput=1
\documentclass[10pt,twocolumn,letterpaper]{article}
\ifreview \usepackage[review]{cvpr} \fi
\ifarxiv \usepackage[pagenumbers]{cvpr} \fi
\ifrebuttal \usepackage[rebuttal]{cvpr} \fi
\ifcamera \usepackage{cvpr} \fi

\usepackage{graphicx}	
\usepackage{amsmath}	
\usepackage{amssymb}	
\usepackage{booktabs}
\usepackage{times}
\usepackage{microtype}
\usepackage{epsfig}
\usepackage{caption}
\usepackage{float}
\usepackage{placeins}
\usepackage{color, colortbl}
\usepackage{stfloats}
\usepackage{enumitem}
\usepackage{tabularx}
\usepackage{xstring}
\usepackage{multirow}
\usepackage{xspace}
\usepackage{url}
\usepackage{subcaption}
\usepackage{xcolor}
\usepackage[hang,flushmargin]{footmisc}

\usepackage[accsupp]{axessibility}

% ...
\definecolor{lblue}{rgb}{0.2, 0.3, 0.8}
\definecolor{purp}{rgb}{0.65, 0.16, 0.65}
\definecolor{red}{rgb}{0.9, 0.1, 0.1}
\newcommand{\LHJ}[1]{{\color{red}{#1}}}
\newcommand{\LHJmodified}[1]{{\color{red}{#1}}}
\newcommand{\hsh}[1]{{\color{purp}{#1}}}

\definecolor{suha_blue}{rgb}{0.1, 0.1, 0.7}
\definecolor{suha_orange}{rgb}{0.9, 0.45, 0.0}

\newcommand{\suhac}[1]{{\color{suha_orange}{(#1)}}}

\newcommand{\Fig}[1]{Fig.~\ref{fig:#1}}
\newcommand{\Sec}[1]{Sec.~\ref{sec:#1}}
\newcommand{\Eq}[1]{Eq.~(\ref{eq:#1})}
\newcommand{\Tbl}[1]{Table~\ref{tab:#1}}

\ifarxiv  \fi

\usepackage{pifont}
\newcommand{\cmark}{\ding{51}}%
\newcommand{\xmark}{\ding{55}}%

\newcommand{\R}[1]{{%
    \textbf{%
        \ifstrequal{#1}{1}{\textcolor{red}{R#1}}{%
        \ifstrequal{#1}{2}{\textcolor{blue}{R#1}}{%
        \ifstrequal{#1}{3}{\textcolor{magenta}{R#1}}{%
        \ifstrequal{#1}{4}{\textcolor{teal}{R#1}}{%
                           \textcolor{cyan}{R#1}%
        }}}}%
    }%
}}

\usepackage{xr-hyper}

\makeatletter
\newcommand*{\addFileDependency}[1]{
  \typeout{(#1)}
  \@addtofilelist{#1}
  \IfFileExists{#1}{}{\typeout{No file #1.}}
}

\makeatother

\definecolor{cvprblue}{rgb}{0.21,0.49,0.74}
\usepackage[pagebackref,breaklinks,colorlinks,citecolor=cvprblue]{hyperref}
\usepackage[capitalize]{cleveref}
\crefname{section}{Sec.}{Secs.}
\crefname{table}{Table}{Tables}
\crefname{figure}{Fig.}{Figs.}

\begin{document}
\title{\paperTitle}
\author{\authorBlock} 
\maketitle
\normalsize

\definecolor{lblue}{rgb}{0.2, 0.3, 0.8}

\begin{abstract}
This paper introduces a novel approach to learning instance segmentation using extreme points, i.e., the topmost, leftmost, bottommost, and rightmost points, of each object.
These points are readily available in the modern bounding box annotation process while offering strong clues for precise segmentation, and thus allows to improve performance at the same annotation cost with box-supervised methods.
Our work considers extreme points as a part of the true instance mask and propagates them to identify potential foreground and background points, which are all together used for training a pseudo label generator.
Then pseudo labels given by the generator are in turn used for supervised learning of our final model.
On three public benchmarks, our method significantly outperforms existing box-supervised methods, further narrowing the gap with its fully supervised counterpart.
In particular, our model generates high-quality masks when a target object is separated into multiple parts, where previous box-supervised methods often fail.

\end{abstract}
\section{Introduction}

Instance segmentation, the task of predicting classes and masks of individual objects at the same time, has been advanced remarkably thanks to supervised learning of deep neural networks~\cite{mask_rcnn, wang2020solo, wang2020solov2, tian2020conditional, cheng2022masked}.
However, it is prohibitively costly to manually annotate a pixel-level mask per instance, which often leads to lack of both class diversity and the amount of training data.
This issue steers the research community towards label-efficient learning approaches such as weakly supervised learning~\cite{ahn2019weakly, zhu2019learning, kim2022beyond, khoreva2017simple, hsu2019weakly, lee2021bbam, tian2021boxinst, lan2021discobox, li2022box, cheng2023boxteacher, li2023sim, Lan_2023_CVPR, cheng2022pointly, tang2022active} and semi-supervised learning~\cite{wang2022noisy, hu2023pseudo, jeong2019consistency, ouali2020semi, liu2022unbiased, zhou2020learning, kim2023devil, rumberger2023actis}.

\begin{figure*}[t]
    \centering
        \includegraphics[width=0.99\textwidth]{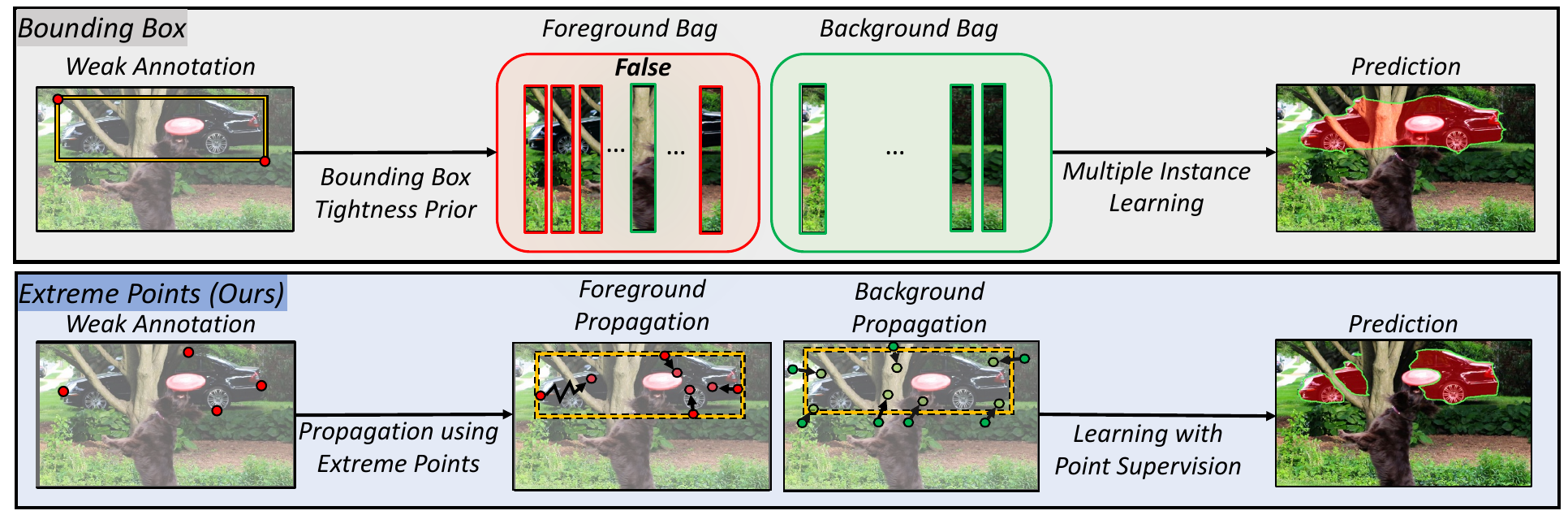}
        \caption{Types of weak supervision and how to utilize it for instance segmentation.
        Top: Box-supervised method relies on bounding box tightness prior, which is often violated by occlusion (foreground bag contains tree trunk).
        As a result, the prediction of the method shows an error in the occluded region.
        Bottom: Extreme point supervised method (Ours) utilizes extreme points as the initial set of foreground points and propagate label through semantic similarity between points. The prediction result demonstrates that our method can predict object mask even in severe occlusion. Best viewed in color.}
    \label{fig:teaser}
    \vspace{-3mm}
\end{figure*} 

Building on this momentum, 
learning instance segmentation using box supervision
has gained considerable attraction recently~\cite{khoreva2017simple, lee2021bbam, hsu2019weakly, tian2021boxinst, lan2021discobox, li2022box, cheng2023boxteacher, li2023sim, Lan_2023_CVPR}.
To train an instance segmentation model with box-supervision, these methods employ a bounding box tightness prior~\cite{hsu2019weakly}, which implies that a vertical (or horizontal) line crossing the bounding box must contain at least one pixel belonging to the object (\Fig{teaser}); this prior has been formulated through various loss functions~\cite{hsu2019weakly, lan2021discobox, cheng2023boxteacher, Lan_2023_CVPR, tian2021boxinst, li2022box}.
Although box-supervision has proved to be effective for learning instance segmentation while keeping annotation costs low, we claim that there is room for further improvement in this direction, particularly due to the fact that it has neglected \emph{extreme points}, a byproduct of the common box annotation process providing a strong clue that helps in estimating the instance mask.

Today, extreme points are freely available in the bounding box annotation process~\cite{kuznetsova2020open}, where human annotators are instructed to click four extreme points of the target object, \ie, topmost, leftmost, bottommost, and rightmost points, rather than to click two corner points of the bounding box.
This is because the former usually ends up requiring less annotation time as the latter often needs to adjust the initial box label multiple times, as demonstrated by Papadopoulos \etal~\cite{papadopoulos2017extreme}.
Moreover, since they are definitely a part of the true mask of the target, extreme points provide a strong clue for segmentation absent in the box supervision. 

\begin{figure}[t]
    \centering
    \includegraphics[width=\linewidth]{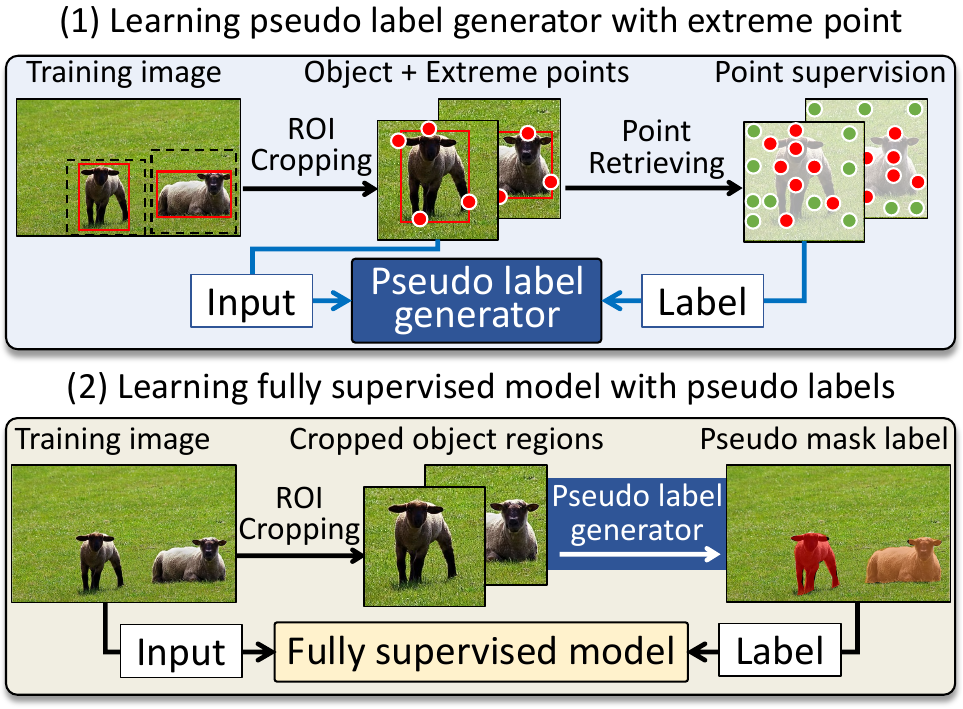}
    \caption{Overview of entire stages of EXITS.
    In the first stage, an image cropped around each object is used as an input to train the pseudo label generator using point-wise supervision, so that the generator learns to predict a binary mask of the object within the cropped image.
    In the second stage, the instance segmentation model learns to detect and segment multiple objects, using the generated pseudo mask labels from the first stage.}
    \label{fig:stage}
    \vspace{-4mm}
\end{figure}

Motivated by this, we study weakly supervised learning for instance segmentation using extreme points to further improve performance without increasing annotation cost. 
Our framework for EXtreme point supervised InsTance Segmentation, dubbed EXITS, considers extreme points as a part of the true instance mask, and exploits them as supervision for training a pseudo label generator.
Then pseudo segmentation labels produced by the generator are in turn used for supervised learning of our final model, which can be any arbitrary networks for instance segmentation.
The overall procedure of EXITS is illustrated in \Fig{stage}. 

The key to the success of EXITS is how to train the pseudo label generator using extreme points. 
A straightforward way is to consider extreme points as foreground and points outside the bounding box as background, and then exploit them for supervised learning. 
However, the pseudo label generator trained in this way fails to generate crisp object masks since most object regions remain unlabeled during training due to the sparsity of extreme points.
To address this issue, EXITS estimates potential foreground and background points within the bounding box by propagating the extreme and background points outside the box.
The propagation process is based on pairwise semantic similarity between points derived by a pretrained transformer encoder so that it reveals 
foreground and background candidates semantically similar with extreme points and nearby background, respectively.
The retrieved points together with the extreme and definite background points serve as supervision for training the pseudo label generator.


As shown in \Fig{teaser}, our pseudo label generator produces high-quality pseudo masks, particularly when a target is divided into multiple parts, and the enhanced quality of pseudo segmentation labels leads to performance improvement of our final model. 
This success is due to the fact that the label propagation is conducted on the fully connected graphs of all the points so that an extreme point can be propagated to spatially distant points.
This alleviates the side-effect of the bounding box tightness prior that is violated in the case of occlusion; the convention box-supervised methods, which rely heavily on the prior, thus often failed in the case.

To quantitatively compare the quality of pseudo labels for separated objects, we measured the pseudo label quality on Separated COCO~\cite{zhan2022tri}, a subset of COCO~\cite{Mscoco} comprising only separated objects.
On the dataset, our method surpassed the previous best method~\cite{Lan_2023_CVPR} by 7.3\%p in mIoU.
We further evaluated EXITS on three public benchmarks, PASCAL VOC~\cite{Pascalvoc}, COCO, and LVIS~\cite{gupta2019lvis}, where EXITS outperformed all the previous box-supervised methods.

In short, the main contribution of this paper is three-fold:
\begin{itemize}
    \item We tackle weakly supervised instance segmentation using extreme points, which can be obtained during bounding box labeling without extra costs.
    \item We introduce a point retrieval algorithm, which effectively leverages extreme points to estimate labels of points in the bounding box.
    Specifically, this algorithm estimates the labels of points based on the probability of propagation to extreme points and background points. 
    \item Our Method achieved the state of the art on three public benchmarks. The qualitative results demonstrated that our method generates high-quality pseudo masks, particularly for separated objects.
\end{itemize}



\label{sec:intro}

\section{Related Work}
\label{sec:relatedwork}

\noindent
\textbf{Instance segmentation.} 
Mask R-CNN~\cite{mask_rcnn} proposes a two-stage approach that first detects regions of interest (RoI) and then predicts segmentation masks within these RoIs.
Subsequent studies have refined this concept by enhancing feature representation~\cite{Liu_2018_CVPR, Chen_2019_CVPR, cai2019cascade} or mask precision~\cite{Huang_2019_CVPR, cheng2020boundary, Zhang_2021_CVPR}.
Then, one-stage methods~\cite{bolya2019yolact,  tian2020conditional, xie2020polarmask, Zhang_2020_CVPR, cheng2022sparse} typically built upon one-stage detectors~\cite{redmon2016you, tian2019fcos} have gained attractions, thanks to their speed and simplicity.
Meanwhile, methods like SOLO~\cite{ wang2020solo} and SOLOv2~\cite{wang2020solov2} introduce box-free one-stage methods without the need for box prediction.
Recently, query-based methods~\cite{dong2021solq, cheng2021per, cheng2022masked, li2022mask, He_2023_CVPR}, inspired by DETR~\cite{carion2020end}, offer impressive performance.
Although these fully supervised methods show remarkable performance, they face practical challenges due to their dependence on costly pixel-wise mask annotation.

\noindent
\textbf{Weakly supervised instance segmentation.} 
Weakly supervised methods using image-level class labels~\cite{zhou2018weakly, ahn2019weakly, zhu2019learning, kim2022beyond}, which depends heavily on class activation maps, have not yet matched fully-supervised performance. Box-supervised methods offer better results with lower annotation costs. The first method in this direction~\cite{khoreva2017simple} refines pseudo masks using GrabCut~\cite{rother2004grabcut}, while recent methods~\cite{tian2021boxinst, lan2021discobox, wang2021weakly, li2022box} incorporate bounding box tightness priors and Multiple Instance Learning (MIL) loss, enhanced with techniques like saliency, color-pairwise affinity, and semantic-correspondence.
Another trend includes the Mask Auto Labeler (MAL)~\cite{Lan_2023_CVPR}, which uses a two-stage process involving pseudo mask generation and model training. Point-based methods~\cite{cheng2022pointly, tang2022active} add point labels to boxes for improved localization. In contrast, our approach leverages extreme points obtained from box annotations for weak supervision, offering robust clues for instance mask estimation.

\noindent \textbf{Extreme point for object annotation.}
An extreme point label is an efficient alternative to a bounding box label, offering a faster annotation process~\cite{papadopoulos2017extreme}.
This approach, being five times quicker than traditional methods, has been increasingly used in object detection training~\cite{kuznetsova2020open, zhou2019bottom} and object segmentation tasks~\cite{papadopoulos2017extreme, maninis2018deep, roth2019weakly, dorent2021inter}. 
DEXTR~\cite{maninis2018deep}, for instance, utilizes extreme points for segmenting arbitrary objects by learning the mapping between input images with extreme points and their segmentation masks. However, DEXTR still requires expensive pixel-level masks for training.
In medical imaging, methods like~\cite{roth2019weakly, dorent2021inter} use extreme points for training voxel segmentation models, generating pseudo-scribble labels by linking extreme points via the shortest path.
Despite these benefits of extreme point label, it has received limited attention in weakly-supervised instance segmentation.
Motivated by this, we introduce to leverage extreme point labels for instance segmentation in diverse scenes predicting precise object masks without using pixel-wise annotations.
Unlike typical approaches in medical imaging that generate scribble pseudo labels based on path-connected object regions, our method uses extreme points to select pseudo-foreground points, which is crucial in scenarios with occlusions, as demonstrated in \Fig{teaser}.

\section{Proposed Method}

\begin{figure*}[t]
    \vspace{-3mm}
    \centering
        \includegraphics[width=0.93\textwidth]{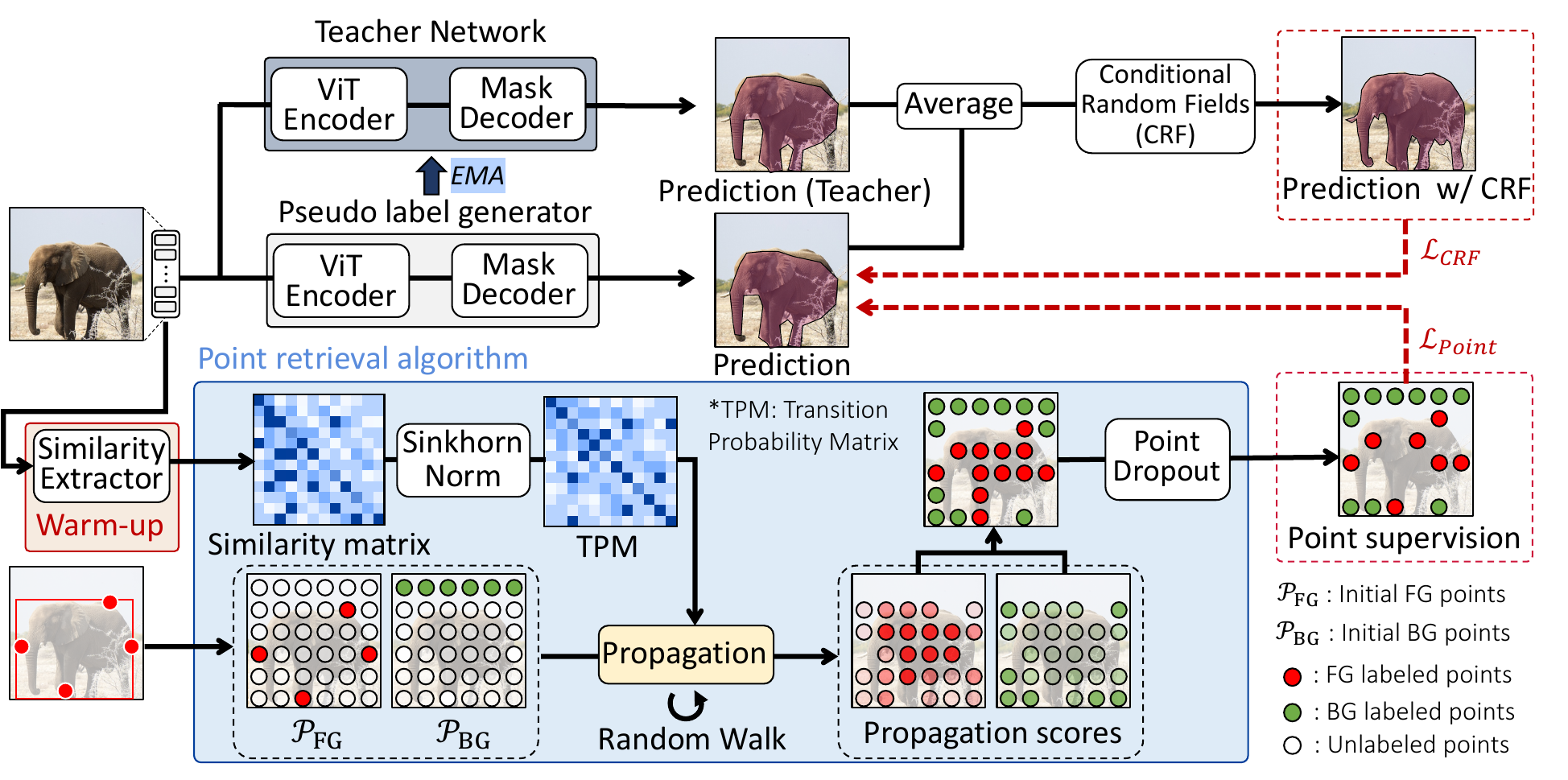}
        \caption{Overview of the first stage of EXITS framework.
                The pseudo label generator is trained on images cropped around each object using the extreme points, aiming to predict binary masks.
                Training leverages two loss functions: $\mathcal{L}_\text{crf}$ aligns images before and after CRF~\cite{crf} processing,
                and $\mathcal{L}_\text{point}$ uses extreme points-derived pseudo point labels for precise pixel-wise supervision.
                To generate these pseudo point labels, EXITS obtains initial foreground and background points from extreme points, then employs the similarity matrix from warm-up trained similarity extractor for label propagation.
                After propagation, pseudo point labels are produced based on the difference of propagation score from the inital foreground and background points.
                Point dropout is applied as an augmentation generating the final pseudo point labels.}
    \label{fig:pipeline}
    \vspace{-4mm}
\end{figure*} 

EXITS consists of two stages: (1) learning a model that generates pseudo segmentation labels of training images using their extreme point labels, and (2) training an instance segmentation model using the pseudo labels.
In the first stage, an object image cropped around each object using extreme points is used as an input to the pseudo label generator so that the model learns to predict a binary mask of the object within the cropped image.
On the other hand, the instance segmentation model in the second stage, which is our final model, learns to detect and segment multiple objects. 
Note that the pseudo label generator deals with an easier task, \ie, instance segmentation on a single object image, which enables to improve the quality of pseudo labels it generates.
The entire pipeline of EXITS is illustrated in~\Fig{stage}.

Since the second stage is the conventional supervised learning that can be applied to any instance segmentation model, this section elaborates mostly on the first stage, in particular, how EXITS provides the pseudo label generator with effective supervision learning for segmentation. 
The overall pipeline of the first stage is illustrated in Fig.~\ref{fig:pipeline}.
The key idea of EXITS is to retrieve pixels likely to belong to the object given the extreme points, and exploit them as supervision for the pseudo label generator.
This idea is realized by propagating the extreme points to other pixels within the input object image, while considering the extreme points as a subset of true pixels of the object.

The remainder of this section first discusses extreme points and advantages of using them (\Sec{subsection1}), and then presents details of the pseudo label generator (\Sec{subsection2}) and the second stage (\Sec{subsection3}) of EXITS.
\subsection{Motivation for Using Extreme Points}
\label{sec:subsection1}

Extreme points are defined as the outermost pixels on an object along the cardinal directions: the topmost point $(x^{(t)}, y^{(t)})$, the leftmost point $(x^{(l)}, y^{(l)})$, the bottommost point $(x^{(b)}, y^{(b)})$, rightmost point $(x^{(r)}, y^{(r)})$. Papadopoulos \etal \cite{papadopoulos2017extreme} demonstrated labeling these points is a more efficient way to bounding box annotation compared to the conventional method of labeling the top-left $(x^{(l)}, y^{(t)})$ and bottom-right $(x^{(r)}, y^{(b)})$ corner points of a box.
This is because such corner points are hard to be identified as they usually do not belong to the target object area, and thus human annotators often have to adjust their initial corner point labels several times.
On the other hand, extreme points can be effortlessly marked and directly converted to a bounding box.
Furthermore, they inherently provide more information for the shape and appearance of the target object than corner points since they lie on the object boundary.

\subsection{Learning Pseudo Label Generator}
\label{sec:subsection2}

The pseudo label generator aims to predict a binary mask of an object given an image cropped around it.
It consists of a vision transformer (ViT) encoder and a mask decoder.
We retrieve points likely to belong to the object (\ie, foreground) or the background, and train the generator using the retrieved points together with the extreme points and definite background points outside the box as supervision.

To be specific, 
the initial set of foreground points is derived from the extreme points as $\mathcal{P}_{\textrm{FG}}:= \big\{ (x^{(t)}, y^{(t)}-\delta), (x^{(l)}+\delta, y^{(l)}), (x^{(b)}, y^{(b)}+\delta), (x^{(r)}-\delta, y^{(r)}) \big\}$, where $\delta$ is a small margin introduced to push the extreme points toward the center of the object so that the points in $\mathcal{P}_{\textrm{FG}}$ are more inward and represent the object more reliably.
On the other hand, the initial set of background points $\mathcal{P}_{\textrm{BG}}$ consists of points located outside the bounding box defined by the extreme points.
To assign pseudo labels to unlabeled points within the bounding box, denoted as $\mathcal{P}_{\textrm{Box}}$, the initial labels from $\mathcal{P}_{\textrm{FG}}$ and $\mathcal{P}_{\textrm{BG}}$ are propagated to them via random walk~\cite{lovasz1993random} with a transition probability matrix, \ie, a matrix of pairwise semantic similarity between points in the input image.
In detail, points in $\mathcal{P}_{\textrm{Box}}$ that are highly likely to be propagated from those in $\mathcal{P}_{\textrm{FG}}$ but not from those $\mathcal{P}_{\textrm{BG}}$ are considered as pseudo foreground. 
Conversely, points in $\mathcal{P}_{\textrm{Box}}$ that are more likely to be propagated from $\mathcal{P}_{\textrm{BG}}$ than $\mathcal{P}_{\textrm{FG}}$ are considered as pseudo background.

\subsubsection{Constructing Transition Probability Matrix}
\label{sec:subsubsection1}

To capture the semantic similarity between points, EXITS leverages an attention matrix obtained from a multi-head self-attention (MHSA) of a ViT encoder.
Since the attention matrix of a randomly initialized or ImageNet-pretrained ViT is not capable of discriminating between foreground and background, we warm-up an extra pretrained ViT encoder, called \textit{similarity extractor}, that is additionally trained for only a few epochs on the target dataset with the multiple instance learning (MIL) loss~\cite{hsu2019weakly, tian2021boxinst}; the loss is defined as
\begin{equation}
    \label{eq:mil_loss}
    \begin{aligned}
        \mathcal{L}_{\text{mil}} = \mathcal{L}_{\text{dice}}\big(&\text{Proj}_{x}(\mathbf{M}), \text{Proj}_{x}(\hat{\mathbf{Y}}_\text{box})\big) \\
        &+ \mathcal{L}_{\text{dice}}\big( \text{Proj}_y(\mathbf{M}), \text{Proj}_y(\hat{\mathbf{Y}}_\text{box}) \big) \;,
    \end{aligned}
\end{equation}
where $\mathbf{M}\in [0,1]^{H\times W}$ is a mask prediction,  $\hat{\mathbf{Y}}_\text{box} \in \{0,1\}^{H\times W}$ is the area of the bounding box, $\mathcal{L}_{\text{dice}}$ indicates the dice loss~\cite{sudre2017generalised}, and $\text{Proj}_x: \mathbb{R}^{H\times W} \mapsto \mathbb{R}^{W}$ and $\text{Proj}_y: \mathbb{R}^{H\times W} \mapsto \mathbb{R}^{H}$ are projection operations that apply the max operation across each column and each row vector of the input matrix, respectively.
Once trained, the similarity extractor is frozen and used to compute the transition probability matrix during training of the pseudo label generator.

We treat each point as a node in a fully connected graph and construct the transition probability between these nodes using their semantic similarity.
To compute the transition probability matrix, a cropped image is divided into $N \times N$ patches and flattened, then fed into the similarity extractor.
The similarity matrix $\mathbf{S} \in \mathbb{R}^{N^2 \times N^2}$ is then derived by averaging the self-attention matrices from multiple attention heads of a transformer layer.
To construct transition probability matrix $\mathbf{T}$ a doubly stochastic form, the Sinkhorn Normalization is applied to $\mathbf{S}$, which is calculated by
\begin{equation}\label{eqn:eq2}
    \mathbf{T} = \frac{\mathbf{A} + \mathbf{A}^\top}{2},   \text{  where  } \mathbf{A} = \text{Sinkhorn}(\mathbf{S}) \;,
\end{equation}
where $\text{Sinkhorn}(\cdot)$ is the Sinkhorn-Knopp algorithm ~\cite{knight2008sinkhorn}.

Building the transition probability matrix using MHSA offers two advantages.
Firstly, since MHSA captures high-level semantic relationship between points, the resulting transition probability matrix prevents points from being propagated to other points with a similar appearance but different semantics.
Secondly, MHSA calculates similarities for all point pairs, thereby naturally yielding a transition probability matrix for a fully connected graph.
This allows the propagation of labels across separated segments of an object, enhancing the accuracy of the label assignment process.

\subsubsection{Generating Pseudo Point Supervision}
\label{sec:subsubsection2}
A pseudo label of $\mathbf{p}_i \in \mathcal{P}_{\text{Box}}$ is assigned by its propagation score calculated by random walk with the transition probability from each member of $\mathcal{P}_{\text{FG}}$ and $\mathcal{P}_{\text{BG}}$ to $\mathbf{p}_i$.
We define the foreground propagation score $\pi_i^{\text{(f)}}$ of $\mathbf{p}_i$ as
\begin{equation}
    \pi_i^{\text{(f)}}
     = \frac{1}{|\mathcal{P}_{\text{FG}}|} \sum_{\mathbf{p}_j \in \mathcal{P}_{\text{FG}}} \mathbf{T}^\alpha(j, i) \;,
    \end{equation}
\noindent where $\mathbf{T}^\alpha(j, i)$ denotes the transition probability that point $\mathbf{p}_j$ propagates to point $\mathbf{p}_i$ through $\alpha$ hops in random walk.
The background propagation score of $\mathbf{p}_i$ is defined in an analogous manner,
\begin{equation}
    \pi_i^{\text{(b)}}
     = \frac{1}{|\mathcal{P}_{\text{BG}}|} \sum_{\mathbf{p}_j \in \mathcal{P}_{\text{BG}}} \mathbf{T}^\alpha(j, i) \;.
\end{equation}
Using these scores, the set of pseudo foreground points $\hat{\mathcal{P}}_{\text{FG}}$ and that of pseudo background points $\hat{\mathcal{P}}_\text{BG}$ are defined as
\begin{equation}
    \begin{aligned}
        &\hat{\mathcal{P}}_\text{FG} := \big\{ (x_i, y_i): \exists (x_i, y_i) \in \mathcal{P}_\text{Box}, \, \pi_i^{\text{(f)}}-\pi_i^{\text{(b)}} \geq \tau_{\text{FG}} \big\} \\
        &\hat{\mathcal{P}}_\text{BG} := \big\{ (x_i, y_i): \exists (x_i, y_i) \in \mathcal{P}_\text{Box}, \, \pi_i^{\text{(f)}}-\pi_i^{\text{(b)}} \leq \tau_{\text{BG}} \big\} \;,
    \end{aligned}
\end{equation}
where $\tau_\text{FG}$ and $\tau_\text{BG}$ are threshold hyperparameters.

\noindent \textbf{Point dropout.}
To enhance the diversity of the pseudo point supervision and prevent overfitting, we adopt an augmentation technique called \textit{point dropout}.
For each epoch, point dropout independently eliminates a random subset from both $\hat{\mathcal{P}}_\text{FG}$ and $\hat{\mathcal{P}}_\text{BG}$, and the removed subsets are excluded from the training process during that epoch.

\subsubsection{Training Objective}
\label{sec:subsubsection3}
\noindent \textbf{Point loss.}
Let $(x_i, y_i)$ denote the 2D coordinates of point $\mathbf{p}_i$.
We construct sparse binary mask $\hat{\mathbf{Y}}\in \mathbb{R}^{N \times N}$ as follows:
\begin{equation}
    \hat{\mathbf{Y}}(x_i, y_i) = 
        \begin{cases} 
            \mathsf{1} & \text{if } \mathbf{p}_i \in  \mathcal{P}_{\text{FG}} \cup \hat{\mathcal{P}}_\text{FG} \\
            \mathsf{0} & \text{otherwise} \;.
        \end{cases}
\end{equation}

\noindent Furthermore, we construct a masking matrix $ \mathbf{K} \in \mathbb{R}^{N \times N} $, which encodes region with point-supervision as follows:
\begin{equation}
    \mathbf{K}(x_i, y_i) = 
    \begin{cases} 
        \mathsf{1} & \text{if } \mathbf{p}_i \in \mathcal{P}_\text{FG} \cup \hat{\mathcal{P}}_\text{FG} \cup \mathcal{P}_\text{BG} \cup \hat{\mathcal{P}}_\text{BG} \\
        \mathsf{0} & \text{otherwise} \;. 
    \end{cases}
\end{equation}

\noindent We employ the dice loss between $\hat{\mathbf{Y}}$ and the predicted mask probability $\mathbf{M}$.
Prior to computing the loss, $\mathbf{M}$ is downsampled to $\tilde{\mathbf{M}} \in [0,1]^{N\times N}$ to match the size with $\hat{\mathbf{Y}}$. Further, we perform an element-wise multiplication between $\tilde{\mathbf{M}}$ and $\mathbf{K}$ so that the loss signal is applied only to the labeled points.
In cases where none of the points is retrieved with the point retrieval algorithm, \ie, $ | \hat{\mathcal{P}}_\text{FG} \cup \hat{\mathcal{P}}_\text{BG} | = 0 $, we apply the MIL loss in~\Eq{mil_loss} additionally.
The point loss is defined as:
\begin{equation}
        \mathcal{L}_{\text{point}} = \mathcal{L}_{\text{dice}}(\tilde{\mathbf{M}} \odot \mathbf{K},\hat{\mathbf{Y}}) + \mathds{1}_{\{|\hat{\mathcal{P}}_\text{FG} \cup \hat{\mathcal{P}}_\text{BG}| = 0\}} \lambda_{\text{mil}} \, \mathcal{L}_{\text{mil}} \;,
\end{equation}
\noindent where $\odot$ is harmard-product operator, $\mathds{1}$ is indicator function, and $\lambda_{\text{mil}}$ is a balancing hyper-parameter.

\noindent \textbf{Conditional random field loss.} 
To further refine the predicted mask, EXITS employs CRF loss as in~\cite{Lan_2023_CVPR}.
Specifically, EXITS utilizes a teacher network obtained by the exponential moving average of training network, \ie, ViT encoder and mask decoder in pseudo labeled generator parameters.
Subsequently, mask predictions from both the training network and the teacher network are averaged to obtain $\mathbf{M}^{\text{avg}}$.
Then, $\mathbf{M}^{\text{avg}}$ is refined through CRF~\cite{crf} by using the mean-field algorithm~\cite{krahenbuhl2011efficient} and utilized as pseudo ground-truth mask using the dice loss as follows:
\begin{equation}
    \begin{aligned}
        \mathcal{L}_{\text{crf}} = \mathcal{L}_{\text{dice}}(\mathbf{M},\textrm{CRF}(\mathbf{M}^{\text{avg}})) \; ,
    \end{aligned}
\end{equation}
where $\textrm{CRF}(\cdot)$ is the CRF operation.
This approach enables the network to yield a more detailed object mask progressively.

\noindent In summary, the overall loss function of EXITS is
\begin{equation}
\begin{aligned}
\mathcal{L}_{overall} =  \lambda_{\text{point}} \mathcal{L}_{\text{point}} +  \lambda_{\text{crf}} \mathcal{L}_{\text{crf}} \;,
\end{aligned}
\end{equation}
where $\lambda_{\text{point}}$ and $\lambda_{\text{crf}}$ are balancing hyper-parameters.

\subsection{Learning a Fully Supervised Model}
\label{sec:subsection3}

In the second stage, EXITS employs the trained pseudo label generator to create pseudo mask labels that serve as ground-truth labels for training a fully supervised instance segmentation model. 
To generate the pseudo mask labels, images containing $k$ instances are cropped around the corresponding extreme point annotations and fed into the generator, yielding a pseudo mask per object.
The decoupled design of the instance segmentation and pseudo labeling models allows for our pseudo labels to be seamlessly integrated into any fully supervised instance segmentation model.

\begin{table*}[t]
    \centering
    \resizebox{\textwidth}{!}{
    \addtolength{\tabcolsep}{3pt}
\begin{tabular}{lcllcccc}
    \toprule
    Method &
      ~~~~Sup~~~~ &
      Backbone &
      InstSeg Model &
      \multicolumn{1}{l}{$\text{Mask AP}_\text{val}$} &
      \multicolumn{1}{l}{$\text{Mask AP}_\text{test}$} &
      \multicolumn{1}{l}{$~~(\%)\text{Ret.}_\text{val}~~$} &
      \multicolumn{1}{l}{$~~(\%)\text{Ret.}_\text{test}~~$} \\ \midrule \midrule
    \multicolumn{8}{c}{\textit{fully-supervised methods}} \\ \midrule
    SOLOv2~\cite{wang2020solov2} &
      M &
      ResNet-50 &
      SOLOv2 &
      37.5 &
      38.4 &
      - &
      - \\
    CondInst~\cite{tian2020conditional} &
      M &
      ResNet-50 &
      CondInst &
      - &
      37.7 &
      - &
      - \\
    FastInst~\cite{He_2023_CVPR} &
      M &
      ResNet-50 &
      FastInst &
      - &
      38.6 &
      - &
      - \\
    SOLOv2~\cite{wang2020solov2} &
      M &
      ResNet-101-DCN &
      SOLOv2 &
      41.7 &
      41.8 &
      - &
      - \\
    SOLOv2~\cite{wang2020solov2} &
      M &
      ResNeXt-101-DCN &
      SOLOv2 &
      42.4 &
      42.7 &
      - &
      - \\
    Mask2Former~\cite{cheng2022masked} &
      M &
      Swin-Small~\cite{liu2021swin} &
      Mask2Former &
      46.1 &
      47.0 &
      - &
      - \\  \midrule
    \multicolumn{8}{c}{\textit{weakly-supervised methods}} \\ \midrule
    DiscoBox~\cite{lan2021discobox} &
      B &
      ResNet-50 &
      SOLOv2 &
      30.7 &
      32.0 &
      81.9 &
      83.3 \\
    BoxTeacher~\cite{cheng2023boxteacher} &
      B &
      ResNet-50 &
      CondInst &
      - &
      35.0 &
      - &
      92.8 \\
    MAL~\cite{Lan_2023_CVPR} &
      B &
      ResNet-50 &
      SOLOv2 &
      35.0 &
      35.7 &
      93.3 &
      93.0 \\
    \rowcolor[HTML]{EFEFEF}
    EXITS (Ours) &
      E &
      ResNet-50 &
      SOLOv2 &
      \textbf{36.1} &
      \textbf{36.9} &
      \textbf{96.3} &
      \textbf{96.1} \\
    BoxInst~\cite{tian2021boxinst} &
      B &
      ResNet-101-DCN &
      CondInst &
      - &
      35.0 &
      - &
      - \\
    DiscoBox~\cite{lan2021discobox} &
      B &
      ResNet-101-DCN &
      SOLOv2 &
      35.3 &
      35.8 &
      84.7 &
      85.9 \\
    BoxLevelSet~\cite{li2022box} &
      B &
      ResNet-101-DCN &
      SOLOv2 &
      35.0 &
      35.4 &
      83.9 &
      83.5 \\
    BoxTeacher~\cite{cheng2023boxteacher} &
      B &
      ResNet-101-DCN &
      CondInst &
      - &
      37.6 &
      - &
      - \\
    SIM~\cite{li2023sim} &
      B &
      ResNet-101-DCN &
      CondInst &
      - &
      37.4 &
      - &
      - \\
    MAL~\cite{Lan_2023_CVPR} &
      B &
      ResNet-101-DCN &
      SOLOv2 &
      38.2 &
      38.7 &
      91.6 &
      92.6 \\
    \rowcolor[HTML]{EFEFEF}
    EXITS (Ours) &
      E &
      ResNet-101-DCN &
      SOLOv2 &
      \textbf{39.8} &
      \textbf{40.2} &
      \textbf{95.4} &
      \textbf{96.2} \\
    DiscoBox~\cite{lan2021discobox} &
      B &
      ResNeXt-101-DCN &
      SOLOv2 &
      37.3 &
      37.9 &
      88.0 &
      88.8 \\
    MAL~\cite{Lan_2023_CVPR} &
      B &
      ResNeXt-101-DCN &
      SOLOv2 &
      38.9 &
      39.1 &
      91.7 &
      91.6 \\
    \rowcolor[HTML]{EFEFEF}
    EXITS (Ours) &
      E &
      ResNeXt-101-DCN &
      SOLOv2 &
      \textbf{40.5} &
      \textbf{40.9} &
      \textbf{95.5} &
      \textbf{95.8} \\
    MAL~\cite{Lan_2023_CVPR} &
      B &
      Swin-Small~\cite{liu2021swin} &
      Mask2Former &
      43.3 &
      44.1 &
      93.9 &
      93.8 \\
    \rowcolor[HTML]{EFEFEF}
    EXITS (Ours) &
      E &
      Swin-Small~\cite{liu2021swin} &
      Mask2Former &
      \textbf{44.2} &
      \textbf{45.0} &
      \textbf{95.9} &
      \textbf{95.7} \\\bottomrule
    \end{tabular}%
    }
    \caption{Results on COCO \texttt{val2017} and \texttt{test-dev}. We report performance using Mask Average Precision (Mask AP) and Retention rate (Ret, \%).
                  Retention rate is the performance ratio compared to its fully supervised counterpart.
                  Each method is trained with the supervision of either a mask (M), bounding box (B), or extreme points (E).
                  Note that the annotation cost of the bounding box and extreme points are equal.}
    \label{tab:coco}
    \vspace{-3mm}
\end{table*}

\section{Experiments}
\label{sec:experiments}

\subsection{Experimental Setting}
\noindent \textbf{Datasets.}
Our method is evaluated on three instance segmentation datasets: COCO~\cite{Mscoco}, PASCAL VOC~\cite{Pascalvoc}, and LVIS v1.0~\cite{gupta2019lvis}.
We utilize the 2017 version of COCO, which contains 115k images for training, 5k for validation, and 20k for testing across 80 classes.
For PASCAL VOC, we employ the augmented version that includes 10,582 training and 1,449 validation images across 20 semantic classes.
LVIS v1.0 contains 164k images spanning 1200+ categories, and we follow the standard partition for training and validation sets as described in~\cite{gupta2019lvis}.
To obtain extreme point annotations, we follow the protocol described in ExtremeNet~\cite{zhou2019bottom}\footnote{\url{https://github.com/xingyizhou/ExtremeNet}}, which converts mask annotations to extreme point annotations.

\noindent \textbf{Evaluation metric.} 
Following previous work~\cite{li2023sim, cheng2023boxteacher, li2022box} we use coco-style \text{Mask AP} as an evaluation metric.
For COCO and LVIS v1.0 datasets, we additionally report \text{Retention Rate} as in MAL~\cite{Lan_2023_CVPR}, which is the ratio of performance compared to its fully supervised counterpart.
\begin{figure*}[t]
    \centering
        \includegraphics[width=0.95\textwidth]{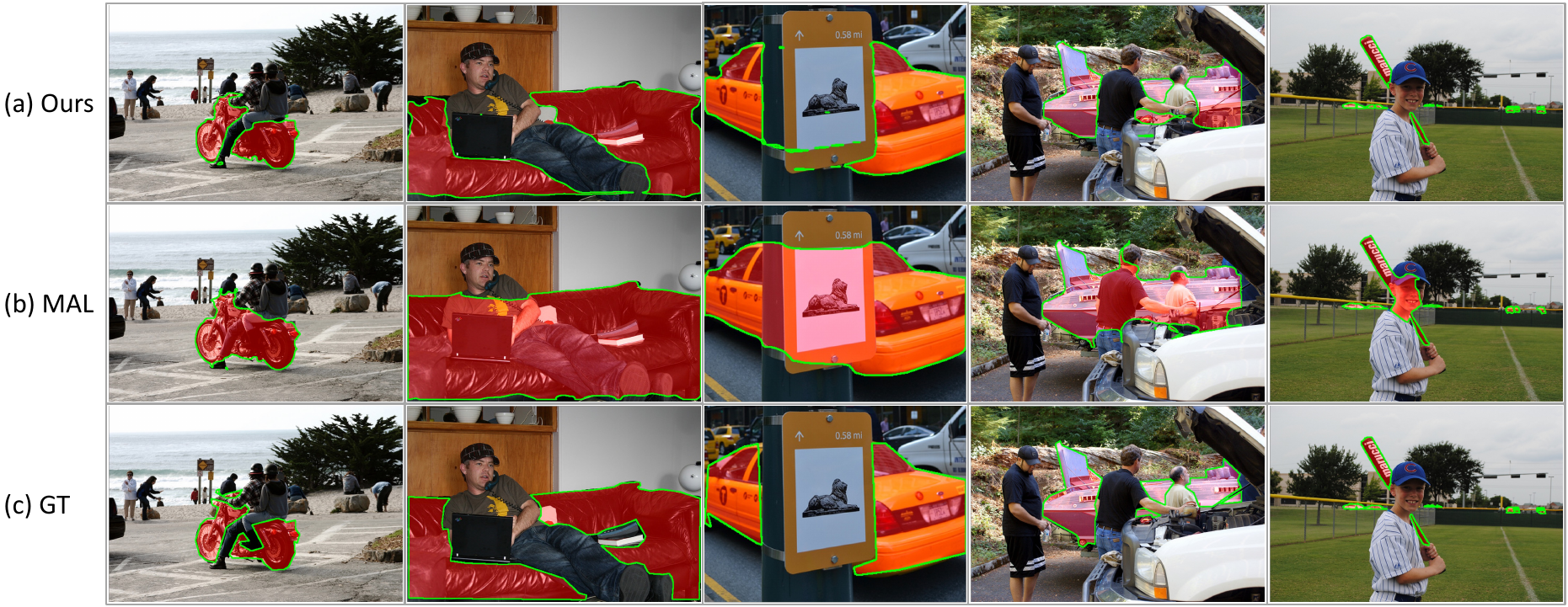}
        \vspace{-2mm}
        \caption{Qualitative comparison of pseudo mask labels on the Separated COCO dataset. (a) Ours, (b) MAL~\cite{Lan_2023_CVPR}, (c) Ground Truth.} 
        \vspace{-3mm}
    \label{fig:pseudomask}
\end{figure*} 
\begin{table}
    \vspace{-3pt}
    \centering
    \renewcommand{\tabcolsep}{3pt}
    \renewcommand\arraystretch{1.1}
    \small
    \begin{tabular}{llccc}
    \toprule
    Method & Backbone & AP & AP$_{50}$ & AP$_{75}$ \\
    \midrule \midrule
    BoxInst~\cite{tian2021boxinst} & ResNet-50 & 34.3 & 59.1& 34.2 \\
    DiscoBox~\cite{lan2021discobox} & ResNet-50 & -  & 59.8  & 35.5 \\
    BoxLevelSet~\cite{li2022box} & ResNet-50 & 36.3  & 64.2  & 35.9\\
    SIM~\cite{li2023sim} & ResNet-50 & 36.7 & 65.5 & 35.6 \\
    BoxTeacher~\cite{cheng2023boxteacher} & ResNet-50 & 38.6  & 66.4  & 38.7 \\
    MAL$^\dagger$~\cite{Lan_2023_CVPR} & ResNet-50 & 37.6 & 64.8& 37.9 \\
    \rowcolor[HTML]{EFEFEF} 
    EXITS (Ours) & ResNet-50 & \textbf{40.4} & \textbf{67.4}  & \textbf{41.4} \\
    \midrule
    BBTP~\cite{hsu2019weakly} & ResNet-101 & -  & 58.9& 21.6 \\ 
    Arun \etal~\cite{arun2020weakly} & ResNet-101 & -  & 57.7  & 31.2 \\   
    BBAM~\cite{lee2021bbam} & ResNet-101 & -  & 63.7  & 31.8 \\
    BoxInst~\cite{tian2021boxinst} & ResNet-101 & 36.4 & 61.4 & 37.0 \\
    DiscoBox~\cite{lan2021discobox} & ResNet-101 & -  & 62.2  & 37.5 \\
    BoxLevelSet~\cite{li2022box} & ResNet-101 & 38.3 & 66.3 & 38.7\\
    SIM~\cite{li2023sim} & ResNet-101 & 38.6 & 67.1& 38.3 \\
    BoxTeacher~\cite{cheng2023boxteacher} & ResNet-101 & 40.3& \textbf{67.8}  & 41.3 \\
    MAL$^\dagger$~\cite{Lan_2023_CVPR} & ResNet-101 & 38.4 & 65.7 & 39.1  \\
    \rowcolor[HTML]{EFEFEF} 
    EXITS (Ours) & ResNet-101 & \textbf{41.4} & 67.7 & \textbf{42.5} \\
    \bottomrule
    \end{tabular}
    \caption{Results on Pascal VOC \texttt{val2012}. Symbol "$\dagger$" denotes the re-implemented results.}
    \label{tab:voc}
    \vspace{-5mm}
\end{table}

\noindent \textbf{Implementation details (first stage).}
We followed the architecture of MAL~\cite{Lan_2023_CVPR} for consistent comparison.
The Standard ViT-Base~\cite{dosovitskiy2020image}, pretrained with MAE~\cite{he2022masked}, served as our ViT encoder, paired with an attention-based mask decoder.
The teacher network is derived from the exponential moving average of the model parameters.
We employ AdamW optimizer~\cite{loshchilov2017decoupled} with the learning rate of $1.5 \times 10^{-6}$, adjusted by cosine annealing scheduler.
An input image is cropped around an object and resized to $512 \times 512$, where data augmentation same as MAL is applied.
We use MHSA of the 10th transformer layer of the similarity extractor as similarity matrix to construct the TPM.
We set the iteration $\alpha$ to 3, the point dropout rate to 0.9, $\tau_{\text{FG}}$ to $\num{1e-3}$, and $\tau_{\text{BG}}$ to $\num{-1e-4}$.
For COCO and LVIS v1.0 datasets, the similarity extractor is trained for 1 and 10 epochs, respectively.
For VOC, the similarity extractor and the pseudo label generator are trained for 8 epochs and 80 epochs, respectively.
More details are given in the supplementary materials.

\noindent \textbf{Implementation details (second stage).}
Various backbone networks and instance segmentation models are adopted for the second stage.
For COCO dataset, we employ ResNets~\cite{resnet}, ResNeXts~\cite{xie2017aggregated}, Swin Transformer~\cite{liu2021swin} as backbone and SOLOv2~\cite{wang2020solov2} and Mask2Former~\cite{cheng2022masked} as instance segmentation model. 
For VOC dataset, we employ the ResNet backbone and SOLOv2 instance segmentation model. 
For LVIS v1.0, we employ ResNeXts backbone and Mask R-CNN~\cite{mask_rcnn} instance segmentation model.
We follow the training configuration of \texttt{mmdetection}~\cite{chen2019mmdetection}\footnote{\url{https://github.com/open-mmlab/mmdetection}}.

\subsection{Comparisons with State-of-the-art}

\noindent \textbf{Results on COCO.}
In~\Tbl{coco}, we compare the performance of EXITS with the baselines trained with the supervision of either a mask (M), bounding box (B), or extreme points (E), on the COCO dataset.
Note that the extreme point has the same labeling cost as the bounding box.
EXITS outperforms the box-supervised baselines in every setting across all the compared backbones and instance segmentation models, indicating that EXITS produces high-quality pseudo labels regardless of the backbone or the applied instance segmentation model.
Especially with the ResNet-101-DCN backbone, EXITS outperforms the state of the arts such as BoxTeacher(+2.6 AP), SIM(+2.8 AP), and MAL(+1.5 AP) by a significant margin.
While the baseline method already achieved a retention rate of over 91\%, EXITS further narrows the performance gap with its fully-supervised counterparts.

\noindent \textbf{Results on PASCAL VOC.}
In~\Tbl{voc}, we compare the performance of EXITS with the baselines on the PASCAL VOC dataset.
EXITS outperforms the box-supervised baselines with both the ResNet50 and the ResNet101 backbones.
Especially with ResNet50 backbone, EXITS shows a significant improvement of 1.8\%p, compared to the previous arts.
This shows that EXITS predicts higher-quality masks for instance segmentation compared to box-supervised methods.

\noindent \textbf{Results on LVISv1.0.}
In~\Tbl{lvis}, we compare the performance of EXITS with MAL~\cite{Lan_2023_CVPR} on the LVIS v1.0 dataset.
EXTIS clearly outperforms MAL in both AP and Ret, which indicates the effectiveness of utilizing extreme points.

\begin{table}[t]
    \centering
    \renewcommand{\tabcolsep}{3pt}
    \renewcommand\arraystretch{1.1}
    \small
        \begin{tabular}{lclcc}
        \hline
        Method & Sup & Backbone & \multicolumn{1}{l}{$\text{Mask AP}_\text{val}$} & \multicolumn{1}{l}{$(\%)\text{Ret.}_\text{val}$} \\ \hline \hline
        \multicolumn{5}{c}{\textit{fully-supervised methods}}                                   \\ \hline
        Mask R-CNN~\cite{mask_rcnn} & M & RNeXt101-32 & 25.5          & -             \\
        Mask R-CNN~\cite{mask_rcnn} & M & RNeXt101-64 & 25.8          & -             \\ \hline
        \multicolumn{5}{c}{\textit{weakly-supervised methods}}                                  \\ \hline
        MAL~\cite{Lan_2023_CVPR}         & B & RNeXt101-32 & 23.7          & 92.9          \\
        \rowcolor[HTML]{EFEFEF} 
        EXITS (Ours)                        & E & RNeXt101-32 & \textbf{24.1}            & \textbf{94.5}            \\
        MAL~\cite{Lan_2023_CVPR}           & B & RNeXt101-64 & 24.5 & 95.0           \\
        \rowcolor[HTML]{EFEFEF} 
        EXITS (Ours)                        & E & RNeXt101-64 & \textbf{24.9}           & \textbf{96.5}            \\ \hline
        \end{tabular}%
    \vspace{-2mm}
    \caption{Results on LVIS v1.0. Best results are noted as \textbf{bold}.}
    \label{tab:lvis}
    \vspace{-3mm}
\end{table}

\subsection{Pseudo-label Quality Comparison}
\begin{figure*}[t]
    \centering
        \includegraphics[width=\textwidth]{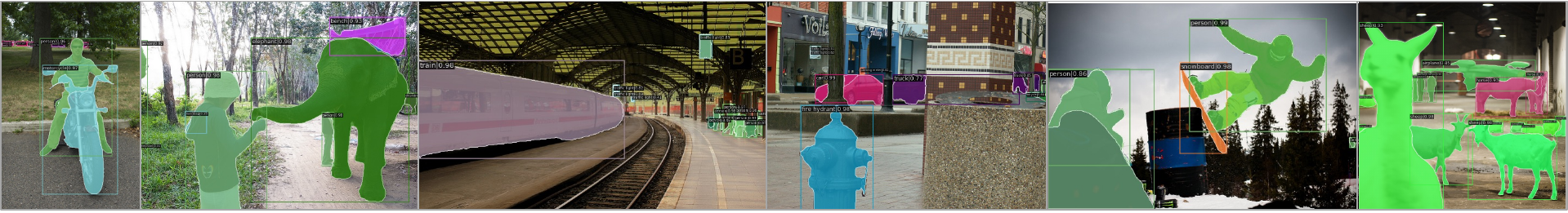}
        \vspace{-6mm}
        \caption{Qualitative results of the final prediction of EXIST on COCO \texttt{test-dev}, using Mask2Former with Swin-Small backbone.
        Our generated pseudo mask labels, EXITS produces high-quality segmentation results, even in separated objects or complex scenes.}
        \vspace{-3mm}
    \label{fig:finalprediction}
\end{figure*}

We evaluate the quality of the generated pseudo mask on COCO and Separated COCO dataset~\cite{zhan2022tri} in mIoU.
Separated COCO is a subset of COCO and consists of objects whose segmentation masks are separated into multiple parts due to occlusion.
In~\Tbl{sepcoco}, we compare the pseudo label quality with MAL~\cite{Lan_2023_CVPR}.
EXITS shows a significant mIoU improvement of 7.3\%p compared to MAL on the Separated COCO dataset, indicating that EXITS generates high-quality masks for separated objects, thanks to its propagation conducted on the fully connected graphs of all points.
This shows that EXITS successfully alleviates the side-effect of the bounding box tightness prior.
In~\Fig{pseudomask}, we conduct a qualitative comparison of pseudo mask labels, where EXITS exhibits superior pseudo label quality compared to MAL.
Thanks to our high-quality pseudo mask labels, the second stage model produces delicate prediction even in separated objects or complex scenes, as illustrated in~\Fig{finalprediction}.

\subsection{Ablation Study}
For the ablation studies, we employ ResNet50 backbone with the SOLOv2 model evaluated on the PASCAL VOC dataset using coco-style AP, AP$_{50}$, AP$_{75}$ metrics. More analysis can be found in the supplement.

\noindent \textbf{Contribution analysis of point set in $\mathcal{L}_{point}$.}
In~\Tbl{component_final}, we evaluate the contributions of the initially labeled point set $\mathcal{P}_{\textrm{FG}} \cup \mathcal{P}_{\textrm{BG}}$, and the pseudo labeled point set $\hat{\mathcal{P}}_{\textrm{FG}} \cup \hat{\mathcal{P}}_{\textrm{BG}}$, when training with $\mathcal{L}_{\textrm{point}}$.
We consider MAL~\cite{Lan_2023_CVPR} as a strong baseline without any point supervision (the first-row of~\Tbl{component_final}).
The improvement from utilizing $\mathcal{P}_{\textrm{FG}} \cup \mathcal{P}_{\textrm{BG}}$ is marginal, showing that using extreme points na\"{i}vely is insufficient to utilize their information for segmentation.
Pseudo point supervision from $\hat{\mathcal{P}}_{\textrm{FG}} \cup \hat{\mathcal{P}}_{\textrm{BG}}$ gives significant performance improvement of 2.4\%p AP, indicating that our point retrieval algorithm is effective.

\noindent \textbf{Effect of point dropout.}
In~\Tbl{dropout}, we show the effectiveness of our point dropout strategy, which leads to 0.6\%p improvement in AP.

\begin{figure}[t]
    \centering
    \includegraphics[width=0.95\linewidth]{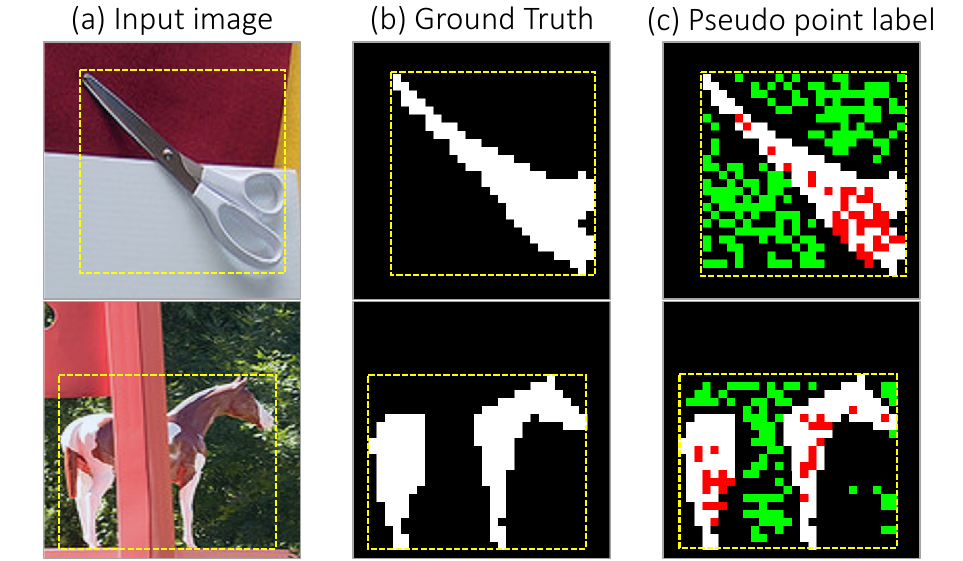}
    \vspace{-2mm}
    \caption{Visualization of pseudo point labels.
    The white points indicate ground truth, the red indicates $\hat{\mathcal{P}}_{\textrm{FG}}$, and the green points indicates $\hat{\mathcal{P}}_{\textrm{BG}}$. To better visualize pseudo point labels, we use a dropout rate of 0.5 in the illustration.
    Best viewed in color.}
    \label{fig:plbl_point}
    \vspace{-3mm}
\end{figure}

\noindent \textbf{Visualizations of pseudo points labels}.
In~\Fig{plbl_point}, we illustrate the generated pseudo point labels from EXITS.
Our pseudo point label accurately captures the object area, effectively excluding the background region even in the occluded areas of the separated objects.
%
\begin{table}[t]
\centering
\resizebox{0.9\linewidth}{!}{%
\begin{tabular}{lcc}
\toprule
             & COCO (mIoU)         & Separated COCO~\cite{zhan2022tri} (mIoU)         \\ \midrule \midrule
MAL~\cite{Lan_2023_CVPR}          & 79.1                 & 59.3                    \\
\rowcolor[HTML]{EFEFEF} 
EXITS (Ours) & \textbf{79.4}              & \textbf{66.6}                    \\ \bottomrule
\end{tabular}
}
\vspace{-2mm}
\caption{Pseudo label quality of the first stage.}
\vspace{-3mm}
\label{tab:sepcoco}
\end{table}
\begin{table}[t]
\centering
\resizebox{0.8\linewidth}{!}{%
\begin{tabular}{ccccc}
\toprule
 $\mathcal{P}_{\textrm{FG}} \cup \mathcal{P}_{\textrm{BG}}$ & $\hat{\mathcal{P}}_{\textrm{FG}} \cup {\hat{\mathcal{P}}_{\textrm{BG}}}$ & AP & AP$_{50}$ & AP$_{75}$ \\ \midrule \midrule
\xmark                 & \xmark                    & 37.6 & 64.8 & 37.9      \\ 
\cmark                 & \xmark                & 38.0 & 65.3 & 38.6      \\
\cmark                 & \cmark                   & \textbf{40.4} & \textbf{67.4} & \textbf{41.4}       \\\bottomrule
\end{tabular}%
}
\vspace{-2mm}
\caption{Ablation study of the effect of points supervision.}
\vspace{-3mm}
\label{tab:component_final}
\end{table}
\begin{table}[ht]
\centering
\resizebox{0.6\linewidth}{!}{
\begin{tabular}{cccc}
\toprule
w/ Point dropout & AP & AP$_{50}$ & AP$_{75}$ \\ \midrule \midrule
\xmark            & 39.8 & 67.1 & 40.4      \\ 
\cmark            & \textbf{40.4} & \textbf{67.4} & \textbf{41.4}      \\    \bottomrule
\end{tabular}%
}
\vspace{-2mm}
\caption{Effect of the point dropout strategy.}
\label{tab:dropout}
\end{table}
\vspace{-2mm}
%
\section{Conclusion}
\label{sec:conclusion}
We have introduced EXITS, a novel framework for learning instance segmentation using extreme points cost-effectively.
EXITS narrows the gap between weakly supervised instance segmentation and its fully supervised counterparts, showing particular strength in segmented objects in severe occlusion scenarios.
On the other hand, even with the use of extreme points, differentiating between occluded objects of the same class continues to be a challenging task.
Our next agenda is to address this issue by using minimal additional supervision, such as center points.

\vspace{1mm}
{
\noindent \textbf{Acknowledgement.}
~This work was supported by the NRF grant and 
the IITP grant 
funded by Ministry of Science and ICT, Korea
(NRF-2018R1A5A1060031, 
 NRF-2021R1A2C3012728, 
 IITP-2019-0-01906, 
 IITP-2022-0-00926). 
 }
\clearpage
\appendix
\setcounter{table}{0}
\setcounter{figure}{0}
\setcounter{equation}{0}
\renewcommand{\theequation}{a\arabic{equation}}
\renewcommand{\thetable}{a\arabic{table}}
\renewcommand{\thefigure}{a\arabic{figure}}
\maketitlesupplementary

In this supplementary material, we provide the following contents omitted from the main paper due to the space limit.
\begin{itemize}
    \setlength\itemsep{0.5em}
    \item Details of pseudo label generator architecture (Sec.~\ref{appendix:architecture})

    \item More experimental details (Sec.~\ref{appendix:experimental})

    \item Analysis on propagation (Sec.~\ref{appendix:propagation})
    
    \item Impact of hyperparameters (Sec.~\ref{appendix:hyperparameter})
    
    \item Analysis on similarity extractor 
    (Sec.~\ref{appendix:similarityextractor})

    \item Time and memory complexity of EXITS 
    (Sec.~\ref{appendix:timeandmemory})

    \item More qualitative results (Sec.~\ref{appendix:morequal})
    
    \item Limitation of the proposed method (Sec.~\ref{appendix:limitation})
\end{itemize}



\section{Details of Pseudo Label Generator Architecture}\label{appendix:architecture}
The network of pseudo label generator consists of the ViT encoder and the mask decoder. 
The architecture of ViT encoder follows the standard vision transformer design, which consists of 12 transformer layers.
We do not use a class token, only the output features are fed into the mask decoder.
The ViT encoder produces the image features $F \in \mathbb{R}^{N \times N \times D}$  from the cropped input image. 

The mask decoder architecture consists of two heads, a pixel-wise head, and a prototype head, a design inspired by YOLACT~\cite{bolya2019yolact}.
The pixel-wise head comprises four convolutional layers, with bilinear interpolation used to upscale the feature resolution between the second and third convolutional layers.
The feature map $F$ goes through the pixel-wise head and resulting $F_\text{pixel} \in \mathbb{R}^{H \times W \times D/3}$.
The prototype head consists of two fully connected layers with ReLU activation function and $D/3$ hidden dimensions. We use average pooling along spatial dimension of $F$, and it go through the prototype head and resulting $F_\text{proto} \in \mathbb{R}^{D/3}$

We produce mask feature map by inner product between $F_\text{pixel}$ and $F_\text{proto}$, and the mask probability map $\mathbf{M}$ is given by 
\begin{equation}
    \mathbf{M} = \sigma(F_\text{pixel} \; F_\text{proto})
\end{equation}
where $\sigma$ denotes sigmoid function.


\section{More Experimental Details}\label{appendix:experimental}
The hyperparameter $\delta$, which is a small margin to push extreme points toward the center of the object, is set as follows: 24 for COCO~\cite{Mscoco}, 16 for LVIS v1.0~\cite{gupta2019lvis}, and 12 for PASCAL VOC~\cite{Pascalvoc}. Note that we push the extreme points with these margin on the resized image space, which is $512 \times 512$.
The hyperparameters $\lambda_{mil}, \lambda_{point}, \lambda_{crf}$, which balance each loss term, are set as follows: 10, 0.5, 0.5 for COCO and LVIS v1.0, and 10, 0.05, 0.5 for PASCAL VOC. 
Note that MIL loss is applied only to samples where pseudo point supervision within the bounding box could not be provided using the point retrieval algorithm, \ie $ | \hat{\mathcal{P}}_\text{FG} \cup \hat{\mathcal{P}}_\text{BG} | = 0 $. This accounts for only about 7\% of the total images.

\begin{table}[ht]
\centering
\resizebox{0.6\linewidth}{!}{
\begin{tabular}{cccc}
\toprule
Index of layer & AP & AP$_{50}$ & AP$_{75}$ \\ \midrule \midrule
$all$         & 39.5 & 66.7 & 40.4      \\ 
\#8           & 39.6 & 66.8 & 41.5      \\ 
\rowcolor[HTML]{EFEFEF}\#10          & \textbf{40.4} & \textbf{67.4} & 41.4      \\ 
\#12          & 40.4 & 66.8 & \textbf{41.8}      \\ 
\bottomrule
\end{tabular}%
}
\vspace{-2mm}
\caption{Index of the transformer layer used for extracting similarity matrix. $all$ refers to the results obtained by averaging the similarity matrices from all the transformer layers.
The rows with gray background represent the values used in our model.}
\label{tab:sup_TPM}
\vspace{-4mm}
\end{table}

\begin{table}[ht]
\centering
\resizebox{0.4\linewidth}{!}{
\begin{tabular}{cccc}
\toprule
$\alpha$ & AP & AP$_{50}$ & AP$_{75}$ \\ \midrule \midrule
1           & 34.0 & 63.0 & 32.4      \\ 
2           & 36.1 & 64.3 & 35.0      \\ 
\rowcolor[HTML]{EFEFEF}3          & \textbf{40.4} & \textbf{67.4} & \textbf{41.4}      \\ 
4          & 39.8 & 66.4 & 40.0      \\ 
$\infty$      & 39.6 & 66.3 & 40.1      \\ 
\bottomrule
\end{tabular}%
}
\caption{Effect of $\alpha$ in propagation process.
The rows with gray background represent the values used in our model.}
\label{tab:sup_alpha}
\end{table}

\section{Analysis on Propagation}\label{appendix:propagation}

\noindent \textbf{Similarity matrix.} 
We extract the semantic similarity between points from the multi-head self-attention of the transformer in the similarity extractor. 
\Tbl{sup_TPM} shows the impact of using different transformer layers for the extraction of the similarity matrix. 
Since earlier layers easily miss high-level semantics, averaging similarity matrices across all layers does not yield the best results.
Therefore, we empirically choose to use $10^{th}$ layer for extracting the similarity matrix.

\noindent \textbf{Effect of number of hops ($\alpha$).}
Table~\ref{tab:sup_alpha} shows the effect of $\alpha$ in propagation process. when $\alpha = 1$, equivalent to generating pseudo point labels directly from the similarity matrix, there is a substantial drop in performance. This indicates that the propagation process is crucial for generating accurate pseudo point labels.
We also measured the performance when the random walk propagation was continued until convergence after an unlimited number of steps, which is also known as the Absorbing Markov Chain~\cite{jiang2013saliency, yeo2016unsupervised, yeo2017superpixel}. It is calculated by
\begin{equation}
\mathbf{T}^{\infty} = (1 - \beta)(\mathbf{I} - \beta \mathbf{T})^{-1}
\label{eq:converge_random_walk}
\end{equation}
where $\mathbf{I}$ denotes identity matrix and $\beta \in [0,1]$ denotes blending coefficient between propagated scores and initial scores. 
In cases where the random walk process converged, we observed the best performance at $\beta = 0.25$; 
however, it still did not outperform the results obtained after three propagation steps. 
Furthermore, considering the increased computational cost needed to compute~\Eq{converge_random_walk}, we set the optimal value of $\alpha$ to 3.

\begin{figure*}[t]
    \centering
        \includegraphics[width=0.9\textwidth]{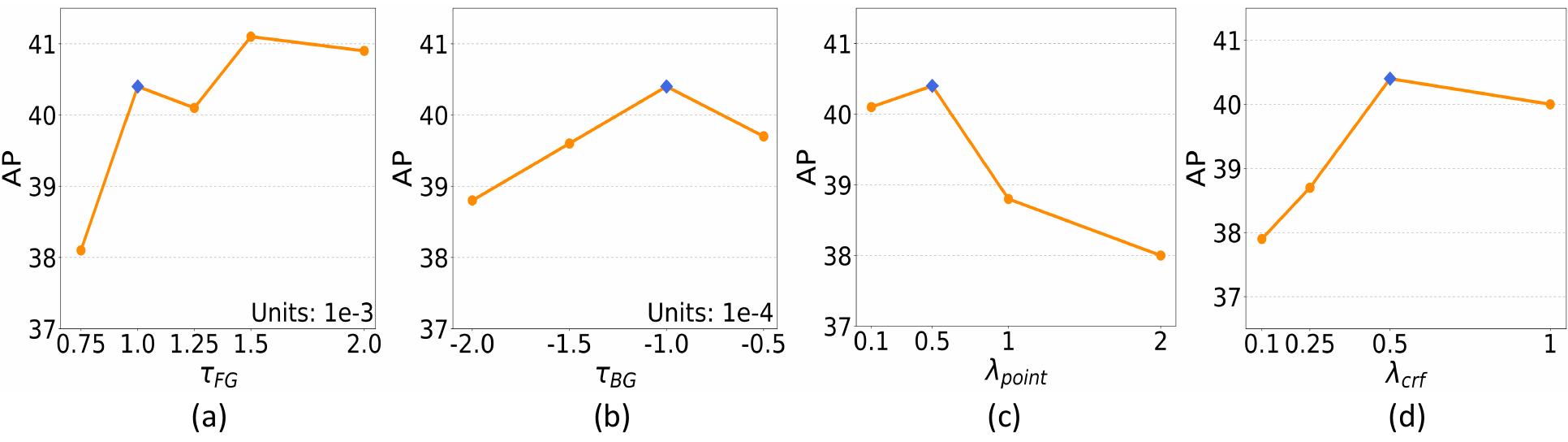}
        \caption{%
        Average Precision (AP) of our second stage model varying hyperparameters.
        The model is evaluated on Pascal VOC using SOLOv2~\cite{wang2020solov2} and ResNet50~\cite{resnet} backbone.
        (a) The foreground point threshold $\tau_\text{FG}$.
        (b) The background point threshold $\tau_\text{BG}$.
        (c) Loss balancing term $\lambda_\text{point}$.
        (d) Loss balancing term $\lambda_\text{crf}$.
        The blue diamond marker indicates the value selected for our final model.} 
    \label{fig:supple_hyper_param}
\end{figure*} 

\section{Impact of Hyperparameters}\label{appendix:hyperparameter}

\noindent \textbf{Effect of }$\tau_\text{FG}$ \textbf{and} $\tau_\text{BG}$\textbf{.}
In \Fig{supple_hyper_param} (a) and (b), we demonstrate the effect of two thresholds, $\tau_\text{FG}$ and $\tau_\text{BG}$, respectively. 
In the case of $\tau_\text{FG}$, we observe that the hyperparameter value we selected are not optimal and there is potential for further performance improvement. 
This indicates that we did not exhaustively tune these parameters using the validation set.

\noindent \textbf{Effect of loss balancing terms.}
In \Fig{supple_hyper_param} (c) and (d), we show the instance segmentation results by using different loss coefficients, $\lambda_\text{point}$ and $\lambda_\text{crf}$. 
Our model demonstrates robustness to these hyperparameter changes, surpassing the baseline~\cite{Lan_2023_CVPR} in every setting.

\section{Analysis on Similarity Extractor}\label{appendix:similarityextractor}

\noindent \textbf{Effect of warm-up training epochs.}\footnote{For an experimental setup, we use the PASCAL VOC and SOLOv2 with ResNet50 backbone as final model. Also we follow $1\times$ schedule for the training configuration of mmdetection.\label{setting}}
As shown in Table \ref{tab:warm-up}, 
more warm-up training leads to better performance by improving background-foreground discrimination of the similarity extractor.

\begin{table}[h]
\centering
\resizebox{\linewidth}{!}{
\begin{tabular}{lccc}
\toprule
 & w/o warm-up  & 4 epochs warm-up  & 8 epochs warm-up (Ours)  \\
\midrule
mask AP & 36.0 & 37.0 & \textbf{37.3} \\
\bottomrule
\end{tabular}
}
\caption{Effect of warm-up training for similarity extractor.}
\label{tab:warm-up}
\vspace{-3mm}
\end{table}

\noindent \textbf{Impact of pretrained weights.}\footref{setting}
In Table \ref{tab:wegiths}, we investigate the impact of pretrained weights for the similarity extractor, as it can have a significant impact on label propagation. In our experiments, using Masked Auto Encoder (MAE)~\cite{he2022masked} pretrained weights shows the best result. We hypothesize that the pixel-wise reconstruction training approach enhances the similarity extractor's ability to learn pixel-level relationships.  

\begin{table}[h]
\centering
\resizebox{0.9\linewidth}{!}{
\begin{tabular}{cccc}
\toprule
Pretrained weights & AP & AP$_{50}$ & AP$_{75}$ \\ \midrule 
\rowcolor[HTML]{EFEFEF}   MAE ~\cite{he2022masked}             & \textbf{37.3} & \textbf{64.4} & \textbf{37.8}      \\ 
ImageNet 22k ~\cite{dosovitskiy2020image}      & 34.2 & 62.3 & 33.3      \\     
ImageNet 1k with DeiT ~\cite{touvron2021training}        & 35.8 & 62.4 & 35.9      \\    
DINO ~\cite{Caron_2021_ICCV}         & 34.7 & 62.0 & 33.9      \\    
\bottomrule
\end{tabular}%
}
\caption{Impact of pretrained weights for similarity extractor.}
\label{tab:wegiths}
\end{table}

\section{Time and Memory Complexity  of EXITS}\label{appendix:timeandmemory}

We compare the training time and number of parameters of EXITS with those of MAL, which is our strong baseline model.
As shown in Table \ref{tab:complexity}, EXITS shows a 20\% increase in training time over MAL due to the warm-up of the similarity extractor and point retrieval process in Stage 1, with an increase in the number of parameters by 86M due to the similarity extractor module.
Although EXITS requires an additional step for warm-up, the consequent increase in training time is only 5\% of the total training time, and the similarity extractor does not affect the space-time complexity of inference.

\begin{table}[h]
\centering
\resizebox{\linewidth}{!}{
\begin{tabular}{c|c|c|c|c|c}
\toprule
\multicolumn{6}{c}{{8 NVIDIA A100 SXM4, COCO dataset, Mask2Former with Swin-Small}} \\ 
\toprule
& & \multicolumn{3}{c|}{Stage 1} & Stage 2 \\ 
\cline{3-6} 
\toprule
& \multirow{2}{*}{Method} & Warm-up & Pseudo label & Pseudo label & Final model  \\
&        & SE & generator training & generation & training \\
\midrule
\multirow{2}{*}{Training time} & MAL~\cite{Lan_2023_CVPR} & - & 23 hrs & 1 hrs 10mins & 71 hrs \\
& EXITS (Ours) & 2 hrs & 26 hrs & 1 hrs 10mins & 71 hrs \\
\midrule
\multirow{2}{*}{\# params} & MAL~\cite{Lan_2023_CVPR} & - & 93M & - & 69M \\
& EXITS (Ours) & 86M & 93M & - & 69M \\
\bottomrule
\end{tabular}}
\caption{Time and memory complexity comparison.}
\label{tab:complexity}
\end{table}

\begin{figure*}[ht]
    \centering
        \includegraphics[width=0.9\textwidth]{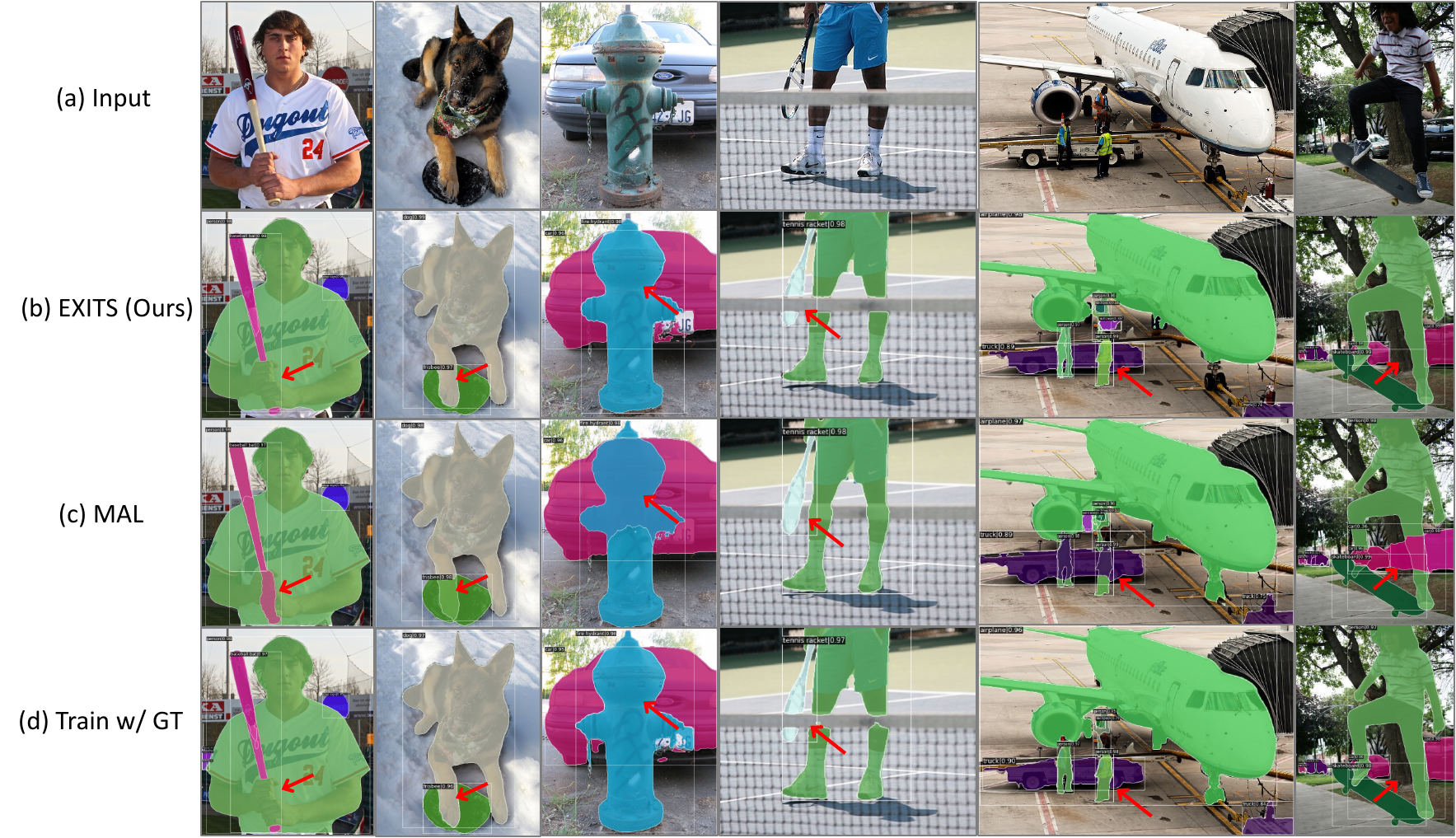}
        \vspace{-3mm}
        \caption{Qualitative comparison of instance segmentation results, especially for separated objects due to occlusions. (a) inputs (b) EXITS (ours), (c) model train with pseudo labels from MAL~\cite{Lan_2023_CVPR}, (d) model train with ground-truth label. The red arrow points to the area where occlusion occurs.} 
        \vspace{3mm}
    \label{fig:supple_phase_2_separated}

    \centering
        \includegraphics[width=0.9\textwidth]{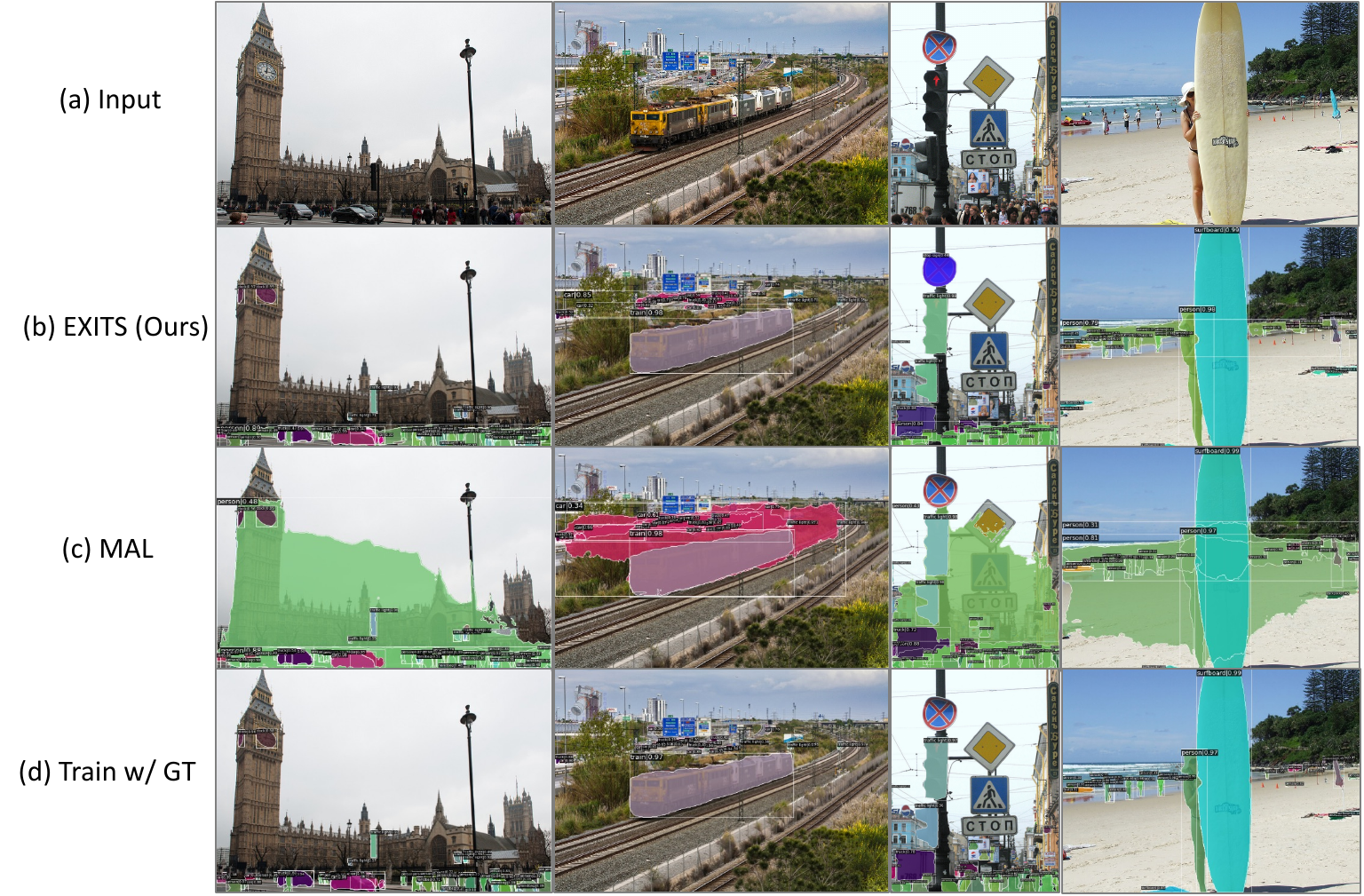}
        \vspace{0mm}
        \caption{Qualitative comparison of instance segmentation results in complex scenes. (a) inputs (b) EXITS (ours), (c) model train with pseudo labels from MAL~\cite{Lan_2023_CVPR}, (d) model train with the ground-truth label.} 
    \label{fig:supple_phase_2_complex}
\end{figure*} 

\section{More Qualitative Results}\label{appendix:morequal}

We visualize the final prediction results produced by Mask2Former~\cite{cheng2022masked} trained with pseudo labels from the pseudo label generator of EXITS using COCO \texttt{test-dev} set.
We also visualized the results of the state-of-the-art box-supervised instance segmentation method, MAL~\cite{Lan_2023_CVPR}, and the upper-bound model trained with the ground-truth labels as a comparison group. 
As can be seen in \Fig{supple_phase_2_separated}, the instance segmentation model trained with our method is capable of generating masks for separated objects, excluding the occluder. This demonstrates almost no difference compared to the results trained with ground-truth labels, while the model trained using pseudo labels generated by MAL struggles in these cases.
Additionally, as illustrated in \Fig{supple_phase_2_complex}, the model trained with our pseudo labels thoroughly predicts even in complex scenes with numerous instances, in contrast to models trained using pseudo labels generated by MAL, which often fail in these scenarios.

\begin{figure*}[t]
    \centering
        \includegraphics[width=0.9\textwidth]{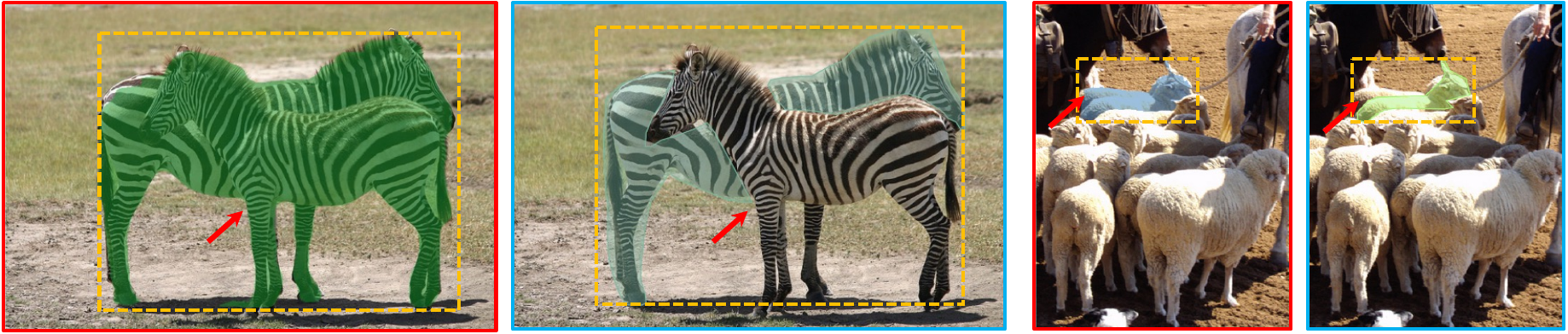}
        \caption{Failure cases of pseudo labels. Our pseudo label generator sometimes fails to predict when instances of the same class are encompassed by the same bounding box. \textcolor{red}{Red} box indicates generated pseudo label from the first stage of EXITS and \textcolor{blue}{blue} box indicates ground-truth label.} 
    \label{fig:supple_failure}
\end{figure*}

\section{Limitation}\label{appendix:limitation}
As observed in \Fig{supple_failure}, our pseudo label generator often mispredicts when multiple objects of the same class are encompassed by the same bounding box.
This issue arises as our point retrieval algorithm assigns pseudo point labels based on the results of the propagation difference between points outside of the bounding box and extreme points. 
One potential clue to solve this issue is to utilize the fact that even objects within the same bounding box have different extreme point annotations. However, this is beyond the scope of this work and will be left for future research.

{\small
\bibliographystyle{ieeenat_fullname}

}

\ifarxiv \clearpage \appendix \section{Appendix Section}
Supplementary material goes here.
 \fi

\end{document}